\documentclass[11pt]{article}
\usepackage{acl2023}
\usepackage{amsmath}
\usepackage{dsfont}
\usepackage{graphicx}
\usepackage{booktabs}
\usepackage{caption}
\usepackage{subcaption}
\usepackage{tikz}
\usetikzlibrary{positioning, arrows.meta, shapes.geometric, fit, backgrounds, calc}
\usepackage{xcolor}
\usepackage{placeins}
\usepackage{float}
\usepackage{listings}
\lstdefinelanguage{sqlx}{
  morekeywords={CREATE, TABLE, VIRTUAL, USING, PRIMARY, KEY, NOT, NULL, DEFAULT, AUTOINCREMENT, FOREIGN, REFERENCES, IF, EXISTS, INTEGER, TEXT, REAL, BLOB},
  morecomment=[l]{--},
  morestring=[b]',
  sensitive=false,
}
\definecolor{aclblue}{rgb}{0.0, 0.0, 0.5}
\usepackage[round,sort&compress]{natbib}
\usepackage[breaklinks,colorlinks=true,linkcolor=aclblue,citecolor=aclblue,urlcolor=aclblue]{hyperref}
\usepackage{url}

\newcommand\blfootnote[1]{%
  \begingroup
  \renewcommand\thefootnote{}\footnote{#1}%
  \addtocounter{footnote}{-1}%
  \endgroup
}

\title{Storage Is Not Memory:\\ A Retrieval-Centered Architecture for Agent Recall}
\author{\textbf{Joshua Adler} \hspace{2em} \textbf{Guy Zehavi} \\
  Sauron Labs \\
  \texttt{josh@sauronlabs.ai}}
\date{}

\begin{document}
\maketitle
\blfootnote{\textit{Technical report. Sauron Labs, 2026.}}

\begin{abstract}
Extraction at ingestion is the wrong primitive for agent memory: content discarded before the query is known cannot be recovered at retrieval time. We propose \textbf{True Memory}, a six-layer architecture that shifts the center of the system from a storage schema to a multi-stage retrieval pipeline operating over events preserved verbatim. The full system runs as a single SQLite file on commodity CPU with no external database, vector index, graph store, or GPU. On LoCoMo \citep{maharana2024}, 1{,}540 questions across 10 multi-session conversations, True Memory Pro reaches 93.0\% accuracy (3-run mean, semantic-match judge; rankings are valid across systems but absolute scores exceed published strict-match baselines) against 61.4\% for Mem0, 65.4\% for Supermemory, and approximately 71\% for Zep \citep{rasmussen2025zep} under a matched \texttt{gpt-4.1-mini} answer model, and 94.5\% for EverMemOS, which uses a GPU-served embedder and a Neo4j graph store. On LongMemEval \citep{wu2024longmemeval}, 500 questions across multi-session conversations, True Memory Pro reaches 87.8\% (3-run mean). On BEAM-1M \citep{tavakoli2026beam}, 700 questions across 35 conversations at the 1-million-token scale, True Memory Pro reaches 76.6\% (3-run mean), above the prior published result of 73.9\% for Hindsight \citep{latimer2025hindsight}. A 56-configuration ablation shows a 1.3-percentage-point spread within the top-performing configuration family. The system also includes an encoding gate (novelty, salience, prediction error) that filters ingestion in production; it is disabled in all reported benchmark numbers because current evaluation instruments reward total recall and cannot score selective ingestion. Gate evaluation is deferred to future work.
\end{abstract}

\section{Introduction}
\label{sec:intro}

Agent memory has emerged as a commercial product category for language models operating across sessions, days, or months, including Mem0 \citep{chhikara2025mem0}, Zep \citep{rasmussen2025zep}, Graphiti \citep{zepai2024graphiti}, Supermemory \citep{supermemory2024}, and EverMemOS \citep{hu2026evermemos}. These systems share a structural pattern in which, at ingestion, a language model parses incoming conversational content into a structured representation, whether atomic facts \citep{chhikara2025mem0}, entity nodes in a knowledge graph \citep{rasmussen2025zep, zepai2024graphiti}, hierarchical summaries \citep{supermemory2024}, or graph-plus-embedding hybrids \citep{hu2026evermemos}, that is then stored in a vector database or graph store and retrieved by similarity at query time.

This pattern inherits the wrong primitive from document search infrastructure, which uses approximate nearest-neighbor indexing \citep{malkov2018hnsw} as exposed by managed or open-source products such as Pinecone, Chroma, Weaviate, and Qdrant \citep{vectordb2026}, where structured records are extracted at ingestion, indexed, and retrieved by similarity. Because extraction runs before the query is known, content discarded at ingestion cannot be recovered at retrieval time.

Retrieval-augmented generation \citep{lewis2020rag} and a body of subsequent work partially mitigate these limitations through query expansion via hypothetical document embeddings \citep{gao2023hyde}, cross-encoder reranking over initial candidates \citep{nogueira2019bertrerank}, and reciprocal rank fusion of lexical and semantic signals \citep{cormack2009rrf}. These refinements improve recall over plain similarity search, but they operate on the structured representation produced at ingest and therefore inherit whatever that step preserved.

We present \textbf{True Memory}, an agent memory architecture that applies an encoding gate at ingestion rather than extracting incoming conversational events into structured records. The gate scores each event for novelty, salience, and prediction error, and events that exceed threshold are preserved verbatim, while higher-order structure such as summaries, entity profiles, and consolidation is computed post-ingestion or deferred to query time. Grounded in reconstructive recall \citep{bartlett1932remembering}, the episodic/semantic distinction \citep{tulving1972episodic}, and levels-of-processing effects on retrievability \citep{craik1972levels}, with the taxonomy of memory distortions from \citet{schacter2001seven} informing our treatment of contradictions, True Memory is implemented as a single SQLite file running on edge compute.

\begin{figure*}[htbp]
\centering
\includegraphics[width=\textwidth]{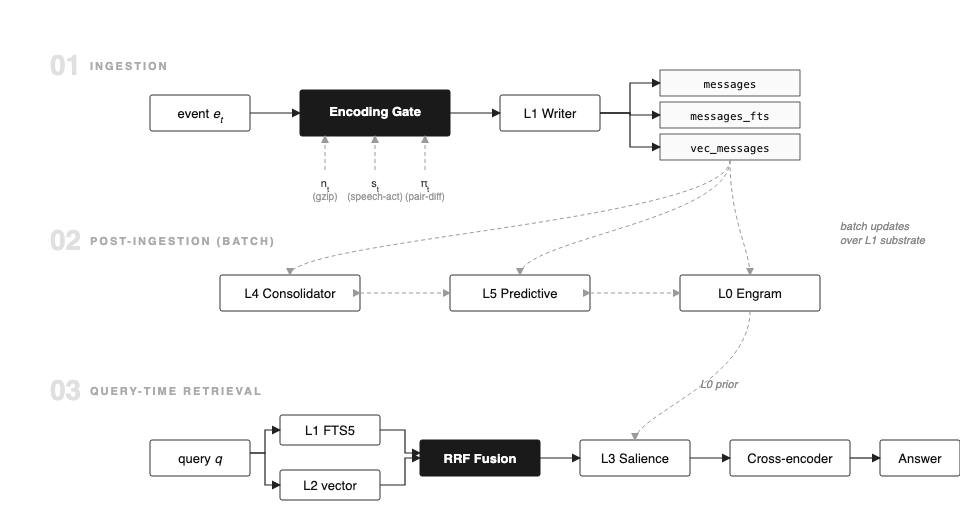}
\caption{True Memory's six-layer architecture across three time-separated phases. \textbf{Ingestion (01):} the encoding gate admits or rejects each event $e_t$ based on novelty ($n_t$), salience ($s_t$), and prediction error ($\pi_t$), computed through independent signal mechanisms; novelty and prediction error query the stored message substrate, while salience scores the event in isolation. Admitted events are written verbatim to \texttt{messages} with corresponding updates to the FTS5 lexical and sqlite-vec dense indices. \textbf{Post-ingestion (02):} batch processes read the L1 event substrate to update L4 consolidation artifacts, the L5 surprise index, and the L0 personality engram. \textbf{Query-time (03):} L1 lexical and L2 dense candidates are fused via reciprocal rank fusion, reweighted by L3 salience conditional on the L0 speaker engram (dashed \emph{prior} arrow), and reranked by a cross-encoder before the top-$k$ are passed to the answer model.}
\label{fig:architecture}
\end{figure*}

The contributions of this paper are fourfold:

\begin{enumerate}
\item \textbf{Architectural reframing.} We formalize agent memory as a retrieval architecture rather than a storage schema (Figure~\ref{fig:architecture}), with encoding, consolidation, and query-time ranking as cooperating stages of the same pipeline rather than independent modules separated by a database.
\item \textbf{Encoding gate.} We introduce a three-signal gated ingestion mechanism (novelty, salience, prediction error) whose novelty and prediction-error signals query the same stored message substrate that the retrieval layers use, while salience scores the incoming event in isolation; ingestion and retrieval therefore share a common representation rather than being separated by a schema boundary.
\item \textbf{Empirical evidence.} We report 93.0\% on LoCoMo (1{,}540 questions, 3-run mean), 87.8\% on LongMemEval (500 questions, 3-run mean), and 76.6\% on BEAM-1M (700 questions at the 1-million-token scale, 3-run mean) under \texttt{gpt-4.1-mini}, trailing only EverMemOS on LoCoMo and exceeding all prior published results on BEAM-1M, with a 56-configuration ablation showing a 1.3-percentage-point spread within the top-performing subfamily.
\item \textbf{Retrieval-as-bottleneck diagnostic.} On 357 LoCoMo questions answered incorrectly by an early True Memory iteration, supplying the full conversation to the same answer model recovered 92\%, locating the bottleneck in the retrieval pipeline rather than the storage layer.
\end{enumerate}

\section{Background}
\label{sec:background}

Vector database infrastructure built on HNSW-class indexing \citep{malkov2018hnsw}, as exposed by products including Pinecone, Chroma, Weaviate, and Qdrant \citep{vectordb2026}, was designed for large-scale document search, where a document is chunked once at indexing time, embedded into a dense vector space, and served by approximate nearest-neighbor lookup. The design is defensible where the corpus is stable, queries fall within a predictable range, and stored items carry roughly equivalent informational weight, since under those conditions the chunk boundaries and the fixed representation are unlikely to matter for most future queries.

Retrieval-augmented generation \citep{lewis2020rag} adapted this infrastructure to language models, and commercial agent memory systems \citep{chhikara2025mem0, rasmussen2025zep, zepai2024graphiti, supermemory2024, hu2026evermemos} extended it to multi-session conversational settings. Query-time refinements accumulated on top, including query expansion via hypothetical document embeddings \citep{gao2023hyde}, cross-encoder reranking over initial candidates \citep{nogueira2019bertrerank}, and reciprocal rank fusion of lexical and semantic signals \citep{cormack2009rrf}; each refinement raises recall over what was preserved, but none of them change what gets preserved.

None of the conditions that justified the document-search design hold in agent memory: the corpus grows with every conversational turn, queries arrive days or weeks after content was stored and frequently ask about details the extractor had no reason to preserve, and informational weight is highly non-uniform, with details mentioned once in passing sometimes serving as the key to a question asked months later.

A parallel line of ML research modifies how language models handle information internally. Memory networks introduce memory slots inside the model's forward pass \citep{sukhbaatar2015memory}, and MemGPT pages content in and out of the model's context window at runtime \citep{packer2023memgpt}. These systems change how a model processes the information it already has, whereas True Memory addresses the prior question of which information reaches the model in the first place. To our knowledge, True Memory is among the first publicly-described agent memory architectures that treat gated ingestion and query-time relevance judgment as cooperating mechanisms, with two of three gate signals querying the same stored message substrate that serves retrieval. The framing draws on reconstructive memory as described by \citet{bartlett1932remembering}, the episodic/semantic distinction due to \citet{tulving1972episodic}, and the complementary-learning-systems account of \citet{mcclelland1995cls}.

Concurrent with this work, several groups have independently explored retrieval-centered architectures for agent memory. These systems share the premise that verbatim event preservation outperforms extraction-based ingestion; they differ in retrieval-pipeline design, storage substrate, and deployment model. In a related direction, \citet{li2026query} propose a query-focused and memory-aware reranker for long-context processing, extending cross-encoder reranking with explicit memory conditioning; their work is complementary to the multi-stage pipeline described here. True Memory's distinguishing characteristic is the single-SQLite-file implementation with zero external dependencies, the three-signal encoding gate, and the emphasis on commodity-hardware deployment.

\section{Design Principles}
\label{sec:principles}

A database returns what was written, whereas a memory returns what is reconstructed at the moment of recall, a distinction foundational in cognitive neuroscience \citep{bartlett1932remembering, schacter2001seven}. Extraction-based agent memory inverts this relationship, committing content to a fixed schema at ingestion and treating retrieval as lookup over that schema, so that the representation becomes the memory. When a query arrives that the schema did not anticipate, the representation cannot accommodate it, and the memory is not available to be retrieved regardless of what was originally said.

A memory system is therefore defined by what it does at the moment of recall rather than by what it wrote down earlier. \citet{tulving1972episodic}'s distinction between episodic and semantic memory is framed at retrieval, where the same underlying substrate serves both and what surfaces depends on how the cue interrogates it, and \citet{craik1972levels} showed that depth of encoding determines later retrievability, coupling encoding and retrieval across time rather than separating them into independent stages.

The scaling argument follows from the same reframing. Language models have finite context windows while conversations extend without bound, so retrieval is the mechanism that lets the bounded window carry the unbounded history. The brain solves the analogous problem through consolidation, by which memories are reorganized over time with emotionally or consequentially salient content preferentially preserved \citep{squire1995retrograde, cahill1995novel}; consolidation on this view is not compression for storage efficiency but preparation for retrieval under cues that cannot be anticipated at encoding. Neither dumping the entire conversation into context nor extracting a lossy summary at ingestion scales to unbounded histories, whereas selective preservation and selective surfacing, decided separately and at different moments, supplies the scalable path.

\section{Architecture}
\label{sec:architecture}

\textbf{True Memory} is organized into six layers (L0--L5) that operate across three time-separated phases: \textit{ingestion}, \textit{post-ingestion batch processing}, and \textit{query-time retrieval}. Stored events are retained in the form they arrived in, and interpretive structure is computed later. The ten pipeline stages below specify each layer by its typed input and output. We adopt the convention, following information-retrieval literature \citep{cormack2009rrf}, that $d$ denotes a candidate item and $q$ denotes a user query; throughout \S\ref{sec:architecture}, $d$ specifically refers to a candidate event, a single admitted message row in the L1 \texttt{messages} table, rather than a document in the traditional document-retrieval sense.

\subsection{Pipeline stages}

The ten pipeline stages below reference the six layers (L0--L5) as follows: Stage 1 uses all three gate signals (L0/L3/L5 derived), Stage 2 = L1 writer, Stage 3 = L4 consolidator, Stage 4 = L5 predictive coder, Stage 5 = L0 engram updater, Stage 6 = L1 lexical retrieval, Stage 7 = L2 dense retrieval, Stage 8 = RRF fusion (cross-layer), Stage 9 = L3 salience reweighter, Stage 10 = cross-encoder reranker.

\subsubsection*{Ingestion phase}

\paragraph{Stage 1: Encoding Gate.}
Input: raw event $e_t$ with text and metadata (sender, timestamp, modality).
Output: binary admit decision plus a signal tag set $\{n_t, s_t, \pi_t\}$ written to the event's metadata on admit.

\enlargethispage{2\baselineskip}
Novelty and salience operate on different axes. Novelty is computed against the stored neighborhood returned by L2 and scores the event relative to what the substrate already contains; salience is computed over the event in isolation and scores properties of the event itself: category, affect, information density. A restatement of a known preference therefore has low novelty but high salience, and a novel passing detail has high novelty but low salience; the gate combines both rather than substituting one for the other.

For each incoming event $e_t$, the gate computes three signals, each through its own mechanism. Novelty and prediction error query the stored message substrate; salience scores the event in isolation:
\newpage
\begin{itemize}
\item \textit{Novelty} ($n_t$) measures how much new information $e_t$ adds to stored memories using gzip compression cost. The gate retrieves the nearest stored messages, concatenates them into a memory text $M$, and computes
\[
n_t = \frac{|\mathrm{gz}(M \,\|\, e_t)| - |\mathrm{gz}(M)|}{|\mathrm{gz}(e_t)|},
\]
where $\mathrm{gz}(\cdot)$ denotes gzip compression at level 6 and $\|\,$ denotes byte concatenation. Redundant information compresses cheaply against the memory context, yielding low $n_t$; novel information is incompressible, yielding high $n_t$. The signal returns $1.0$ when memory is empty and $0.05$ for trivially short messages (compressed size below 10 bytes). This replaced cosine-similarity inversion, which is anti-correlated with novelty in conversational data: noise like ``ok'' is semantically distant from factual memories while important updates are semantically close. Validated in a 120-variant sweep: AUC 0.788 vs.\ 0.484 for cosine baseline. Implementation: \texttt{truememory.ingest.encoding\_gate}.
\item \textit{Salience} ($s_t$) delegates to a hybrid scorer ({\small\texttt{truememory.ingest.encoding\_salience}}) that routes on message length: messages of 50 characters or fewer are scored by a rule-based speech-act classifier that is length-independent, returning fixed scores by linguistic function (e.g., commitments $\to 0.8$, corrections $\to 0.6$, noise $\to 0.02$, questions $\to 0.2$); messages longer than 50 characters are scored by the L3 retrieval salience scorer (\texttt{truememory.\allowbreak{}salience.\allowbreak{}compute\_\allowbreak{}message\_\allowbreak{}salience}), which evaluates length, numbers, dates, and emotional markers.
\item \textit{Prediction error} ($\pi_t$) uses embedding pair-difference scoring. The gate embeds the cross-pair $(e_t \; [\mathrm{SEP}] \; m_1)$ and the self-pair $(m_1 \; [\mathrm{SEP}] \; m_1)$, where $m_1$ is the nearest stored memory, using the same embedding model that serves vector search. Prediction error is
\[
\pi_t = 1 - \cos\!\big(\mathbf{v}_{\mathrm{cross}},\; \mathbf{v}_{\mathrm{self}}\big).
\]
When the cross-pair embedding diverges from the self-pair, the message says something different about the same topic. The signal applies two early-exit filters: messages classified as noise by the salience scorer are assigned $\pi_t = 0$ without embedding, and messages whose nearest stored memory has low topical relevance (below a cosine threshold) are also assigned $\pi_t = 0$. Validated in a 200-variant sweep: AUC 0.730 standalone, gate AUC 0.816 in three-signal combination; independent of novelty ($r = 0.30$) and salience ($r = 0.23$). No additional model is required. Implementation: \texttt{truememory.ingest.encoding\_gate}.
\end{itemize}

The gate applies a salience floor $s_{\min} = 0.10$: messages whose salience falls below $s_{\min}$ are rejected regardless of their gate score, preventing high-novelty off-topic noise from passing. For messages above the floor, admit $e_t$ iff
\begin{equation}
\frac{\lambda_n \cdot n_t + \lambda_s \cdot s_t + \lambda_\pi \cdot \pi_t}{\lambda_n + \lambda_s + \lambda_\pi} \;\geq\; \tau + \Delta_c, \label{eq:gate}
\end{equation}
where $\Delta_c$ is a per-category threshold offset (correction $-0.06$, decision $-0.04$, relationship $-0.04$, all others $0$) that lowers the bar for high-value categories. The weights $(\lambda_n, \lambda_s, \lambda_\pi, \tau) = (0.25, 0.20, 0.30, 0.30)$ are production defaults. The numerator is normalized by the weight sum so the score lands in $[0, 1]$ regardless of weight magnitudes.

In the benchmark configuration used throughout \S\ref{sec:results}, the gate is disabled and every input event is admitted ($\tau = -\infty$).

\paragraph{Stage 2: L1 Writer.}
Input: admitted event plus metadata and signal tag set.
Output: one row in \texttt{messages}, a corresponding update to the \texttt{messages\_fts} FTS5 index, and a 256-dimensional vector row in \texttt{vec\_messages} (sqlite-vec).

\subsubsection*{Post-ingestion phase (batch)}

\paragraph{Stage 3: L4 Consolidator.}\footnote{Inspired by memory consolidation as described in \citet{squire1995retrograde}.}
Input: event clusters over a sliding window of recent \texttt{messages} rows.
Output: summary rows (one per cluster), contradiction records (pairs of rows whose claims conflict, indexed by entity and predicate), and timeline rows (time-ordered assertions with superseded-by links).

\paragraph{Stage 4: L5 Predictive Coder.}\footnote{Inspired by the predictive-coding framework of \citet{rao1999predictive}.}
Input: the event stream in temporal order.
Output: a per-event surprise score $\sigma_t \in [0, 1]$ stored in \texttt{surprise\_scores} and used as a retrieval boost. In the production implementation, $\sigma_t$ is a weighted combination of the fraction of newly-extracted fact fingerprints (numbers, proper nouns, dates, event keywords, definitional predicates) not yet present in the accumulated fact set, plus small length, detail, and event bonuses, with explicit contradiction detection for update verbs. There is no learned predictor $\hat{v}_t$ in the current implementation; the surprise signal is computed directly over extracted fact sets rather than as an L2 residual.

\paragraph{Stage 5: L0 Engram Updater.}\footnote{Inspired by the theory-of-mind framework of \citet{gallagher2003theory}.}
Input: full event history for a given speaker.
Output: a speaker-preference attribute map in \texttt{entity\_profiles} (one row per distinct entity), plus a 256-dimensional character-n-gram style vector in \texttt{entity\_style\_vectors} (hashed char-(3,4,5)-grams, L2-normalized, mean-pooled across a speaker's messages), both refreshed on schedule. At query time the style vector is used as a retrieval prior, scoring candidate messages by cosine similarity to the queried entity's writing profile.

\subsubsection*{Query-time phase}

\paragraph{Stage 6: L1 Lexical Retrieval.}
Input: query $q$.
Output: candidate set $C_1 = $ top-$k$ FTS5 matches over \texttt{messages\_fts} (rank by BM25).

\paragraph{Stage 7: L2 Dense Retrieval.}
Input: query $q$, embedded to $v_q$.
Output: candidate set $C_2 = $ top-$k$ cosine-nearest rows from \texttt{vec\_messages}.

\paragraph{Stage 8: RRF Fusion.}
Input: $(C_1, C_2)$ (and the optional separation list).
Output: fused candidate set $C_{\mathrm{fused}}$ ranked by a per-source-weighted extension of reciprocal rank fusion. The original formulation of \citet{cormack2009rrf} aggregates candidate rankings as
\begin{equation}
\mathrm{RRF}_{\mathrm{Cormack}}(d) = \sum_{r \in R} \frac{1}{k + r(d)}, \qquad k = 60, \label{eq:rrf-cormack}
\end{equation}
where $R$ is the set of contributing rank lists, $r(d)$ is the rank of $d$ in list $r$, and $k = 60$ is the smoothing constant from that work; individual rank lists are not weighted, so the formulation is the uniform case $w_r \equiv 1$. True Memory extends this with a per-source weight $w_r$ to reflect the empirically different contributions of lexical and dense rank lists in multi-session conversational retrieval, and to accommodate the optional separation rank list whose signal is weaker than either primary source:
\begin{equation}
\mathrm{RRF}(d) = \sum_{r \in R} w_r \cdot \frac{1}{k + r(d)}, \qquad k = 60, \label{eq:rrf}
\end{equation}
where $R$ contains the FTS5, dense, and (when present) separation rank lists, with $w_{\mathrm{fts}} = w_{\mathrm{vec}} = 1$ and $w_{\mathrm{sep}} = 0.8 \times w_{\mathrm{vec}}$ in the reference implementation; separation search fires only when the corpus contains more than five distinct senders, because in two- or three-person conversations the sender/recipient prefix produces a uniform ranking that dilutes the primary signals (\texttt{truememory/hybrid.py}). The smoothing constant $k = 60$ is unchanged from \citet{cormack2009rrf}. A principled calibration sweep over the source weights is left for future work.

\paragraph{Stage 9: L3 Salience Reweighter.}\footnote{Inspired by levels-of-processing effects on retrievability \citep{craik1972levels}.}
Input: $C_{\mathrm{fused}}$, plus query-side signals (question type, timestamp) and the L0 speaker engram.
Output: $C_{\mathrm{fused}}$ reweighted by a pipeline of conditional scalar adjustments applied sequentially to each candidate's RRF score:
\begin{equation}
\mathrm{score}_{L_3}(d, q) = \mathrm{RRF}(d) \;\circ\; f_t(d, q) \;\circ\; f_{L_0}(d), \label{eq:l3}
\end{equation}
where $\circ$ denotes sequential application (i.e., each factor either multiplies the running score or injects new candidates) and each factor fires only when its triggering condition is met:
\begin{itemize}
\item $f_t(d, q)$ is a conditional temporal boost: when the query carries temporal intent and the candidate $d$ already appears in the temporally-filtered result set, the running score is multiplied by $1.3$; otherwise $f_t = 1$ (\texttt{truememory/engine.py}).
\item $f_{L_0}(d)$ is a conditional personality-prior injection: when the query has personality intent, L0 profile and style-vector results are appended to the candidate set with scores scaled to $0.8 \times \max(\mathrm{existing\ scores})$ for entity-profile results and $0.9 \times \max$ for style-vector results, so that the L0 prior informs but does not dominate factual retrieval (\texttt{truememory/engine.py}).
\end{itemize}
Results falling below a minimum salience threshold are dropped from the candidate set before reranking. When the \texttt{surprise\_scores} table is populated, candidates with a positive surprise score $\sigma_t > 0$ have their score multiplied by $(1 + \alpha \cdot \sigma_t)$ where $\alpha = 0.2$ (configurable via \texttt{TRUEMEMORY\_\allowbreak{}ALPHA\_\allowbreak{}SURPRISE}); candidates with $\sigma_t = 0$ are left untouched, and the step is a no-op when the surprise index has not been built. Each multiplicative factor is scalar, so $\mathrm{score}_{L_3}(d, q)$ is itself scalar. The pipeline is compositional: factors fire independently and in sequence, with inactive factors contributing $1$ (identity).

\paragraph{Stage 10: Cross-encoder Reranker with Modality Fusion.}
Input: $(q, d.\text{text})$ pairs for the top $N$ items in the reweighted $C_{\mathrm{fused}}$.
Output: scalar relevance score per pair, optionally adjusted by a modality factor $f_m(d, q)$ that examines the candidate's modality and the question type of the query (detail questions penalize summary modalities by $0.7\times$; synthesis questions boost them by $1.2\times$; general questions apply no adjustment), implemented in \texttt{truememory/reranker.py}. The top-$k$ (default $k = 10$) after sorting are passed to the answer model. In the benchmark configuration used throughout \S\ref{sec:results}, the pre-rerank window is 100 (see Table~\ref{tab:harness}).

\section{Implementation}
\label{sec:implementation}

True Memory's entire persistent state is a single SQLite database file, and its runtime dependencies are Python and standard scientific-computing libraries. No external database, vector index, graph store, GPU, or cloud service is required at any configuration.

\paragraph{Storage substrate.} The full storage layer consists of SQLite with the FTS5 full-text-search module \citep{hipp2000sqlite} for lexical search and sqlite-vec \citep{garcia2024sqlitevec} for dense-vector retrieval. Events are stored verbatim in a \texttt{messages} table with metadata (sender, recipient, timestamp, category, modality), while entity profiles, summaries, fact timelines, and surprise indices occupy their own tables populated by post-ingestion batch processes. The database file is portable, backupable as a single file, and inspectable with any SQLite tool.

\subsection{Schema snippet}
\label{sec:schema}

The following is a verbatim excerpt of the production DDL from \texttt{truememory/storage.py} (Figure~\ref{fig:schema}); tables and triggers not central to the exposition are elided.

\noindent\includegraphics[width=\columnwidth]{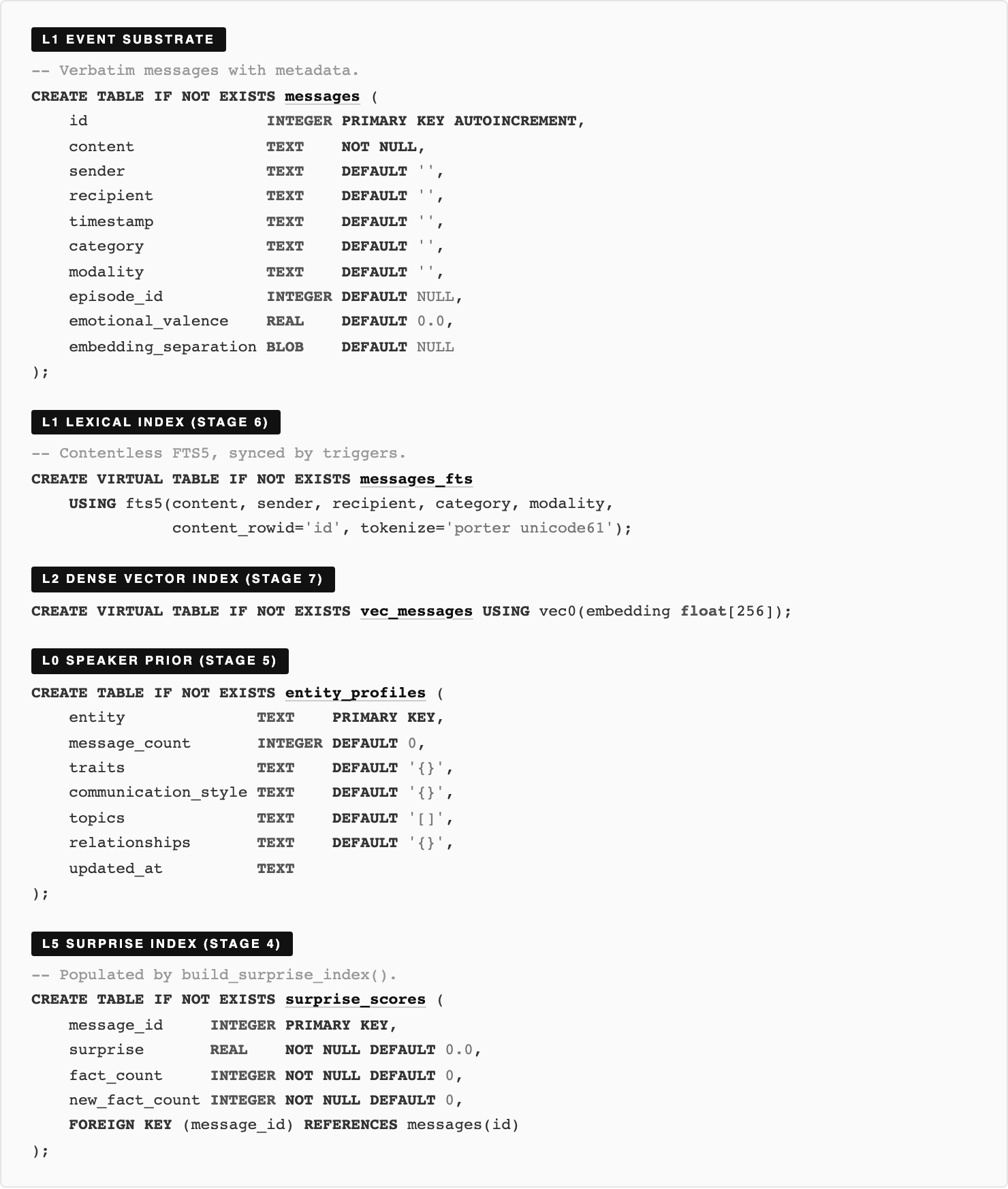}

{\small\sloppy
\captionof{figure}{Production schema excerpt. Each section corresponds to a layer in the architecture of \S\ref{sec:architecture}: \texttt{messages} is the L1 verbatim event substrate, \texttt{messages\_fts} is the L1 FTS5 lexical index, \texttt{vec\_messages} is the L2 dense vector index (created lazily by \texttt{vector\_search.py}), \texttt{entity\_profiles} is the L0 speaker engram, \texttt{entity\_style\_vectors} stores L0 char-n-gram style profiles, and \texttt{surprise\_scores} is the L5 prediction-error index (created lazily by \texttt{predictive.py}). The full DDL also includes \texttt{metadata} (embedder identity and schema version), \texttt{episodes} (6-hour-gap session boundaries), \texttt{landmark\_events} (life events such as job changes and moves), \texttt{causal\_edges} (forward/backward cause chains), and \texttt{entity\_relationships} (Dunbar-hierarchy relationship graph; populated when entity-sheet mode is enabled).}
\label{fig:schema}
\par}

\paragraph{Hardware footprint.} A lightweight configuration uses on the order of 30 million total parameters across the embedder and reranker, yielding approximately 123 MB of static model weights at \texttt{fp32} (approximately 61 MB at \texttt{fp16}), with runtime working-set RAM under 1 GB that is dominated by the Python scientific-computing stack rather than the models themselves; the system runs comfortably on any commodity single-board computer with at least 1 GB of available RAM, and a Raspberry Pi 4 is sufficient for full deployment. Larger configurations use embedders in the hundreds of millions of parameters and approximately 2 to 4 GB of runtime RAM, and none require a GPU. Smaller configurations trade a few points of accuracy for significantly smaller footprint.

\paragraph{Cost profile.} True Memory's infrastructure cost is bounded by local compute, so only the answer-model API call incurs marginal expense; at 100 queries per day this is approximately \$12 per month under \texttt{gpt-4.1-mini} pricing. For reference, EverMemOS is open source and free to self-host, though its pipeline requires a Neo4j graph store, a GPU-served embedder, and multi-service cloud orchestration that carry infrastructure costs not reflected in the benchmark harness. Meanwhile, Mem0, Supermemory, and Zep operate on tier-based pricing models in the \$15 to \$400 per month range depending on tier and usage.

\subsection{Compute and latency budget}
\label{sec:compute}

The three tiers use the following model configurations. Embedder: Model2Vec \texttt{potion-base-8M} \citep{tulkens2024model2vec} (256-dimensional static embedding) for Edge, and Qwen3-Embedding-0.6B \citep{qwen2025embedding} truncated to 256 dimensions via Matryoshka representation learning for Base and Pro. Reranker: \texttt{cross-encoder/\allowbreak{}ms-marco-\allowbreak{}MiniLM-L-6-v2} (22M parameters, CPU-friendly) for Edge, and \texttt{Alibaba-NLP/\allowbreak{}gte-reranker-\allowbreak{}modernbert-base} (149M parameters) for Base and Pro. Pro adds hypothetical-document query expansion \citep{gao2023hyde}. Answer model: \texttt{openai/gpt-4.1-mini} via the OpenAI API across all tiers. A per-stage latency breakdown is left for future work.

\section{Results}
\label{sec:results}

The numbers that follow reflect the retrieval stack alone; the encoding-gate disclaimer from the abstract applies throughout. All TrueMemory results were produced on the v0.6.0 pipeline (\texttt{main} branch). OpenRouter API calls were made between April--May 2026.

We evaluate True Memory on three public benchmarks of long-term conversational memory: LoCoMo \citep{maharana2024}, LongMemEval \citep{wu2024longmemeval}, and BEAM-1M \citep{tavakoli2026beam}.

\paragraph{Evaluation harness.} All three benchmarks are run through the same harness conventions, including deterministic answer and judge generation, \texttt{gpt-4o-mini} as a three-run majority judge, and a capped top-$k$ retrieval window feeding the answer model. All True Memory scores in this section are 3-run means unless otherwise noted; individual run scores and standard deviations are reported where the spread is informative.

\begin{table*}[htbp]
\centering\small
\caption{Evaluation harness settings.}
\label{tab:harness}
\setlength{\tabcolsep}{4pt}
\begin{tabular}{lrrr}
\toprule
Setting & LoCoMo & LongMemEval & BEAM-1M \\
\midrule
Questions ($N$) & 1{,}540 & 500 & 700 \\
Session structure & 10 multi-sess.\ convs & 38--62 haystack sess. & 35 convs at 1M tok. \\
Answer model (TM rows) & \texttt{gpt-4.1-mini} & \texttt{gpt-4.1-mini} & \texttt{gpt-4.1-mini} \\
Judge model & \texttt{gpt-4o-mini} & \texttt{gpt-4o-mini} & \texttt{gpt-4o-mini} \\
Judge runs / question & 3 (majority) & 3 (majority) & 3 (majority) \\
Answer temperature & 0 & 0 & 0 \\
Judge temperature & 0 & 0 & 0 \\
Max answer tokens & 200 & 200 & 200 \\
Retrieval top-$k$ (pre-rerank) & 100 & 100 & 100 \\
\bottomrule
\end{tabular}
\end{table*}

Answer models for competitor rows are reported per row in Table~\ref{tab:locomo} and vary across systems.

\subsection{LoCoMo}
\label{sec:locomo}

\begin{table*}[htbp]
\centering\footnotesize
\setlength{\tabcolsep}{4pt}
\caption{LoCoMo accuracy (v0.6.0, 3-run means for True Memory rows) with 95\% Wilson confidence intervals. $N = 1{,}540$.}
\label{tab:locomo}
\begin{tabular}{lrrrrl}
\toprule
System & Accuracy & 95\% CI & Correct & Cost / correct & vs row above \\
\midrule
EverMemOS$^\ast$ \citep{hu2026evermemos} & 94.48\% & [93.23, 95.51] & 1{,}455 & \$0.0010$^\dagger$ & --- \\
\textbf{TM Pro}$^\ddagger$ & \textbf{93.0\%} & [91.60, 94.16] & 1{,}432 & \$0.0013 & $\chi^2=4.09$, $p{=}0.043$ \\
\textbf{TM Base}$^\ddagger$ & \textbf{92.01\%} & [90.54, 93.26] & 1{,}417 & \$0.0011 & $\chi^2=1.12$, $p{=}0.29$ \\
\textbf{TM Edge}$^\ddagger$ & \textbf{89.65\%} & [88.05, 91.07] & 1{,}381 & \$0.0010 & $\chi^2=6.54$, $p{=}0.011$ \\
RAG (ChromaDB) & 86.17\% & [84.35, 87.80] & 1{,}327 & \$0.0011 & $\chi^2=11.56$, $p{=}6.7{\times}10^{-4}$ \\
Engram & 84.55\% & [82.65, 86.26] & 1{,}302 & \$0.0011 & $\chi^2=2.28$, $p{=}0.13$ \\
BM25 & 80.45\% & [78.40, 82.36] & 1{,}239 & \$0.0011 & $\chi^2=17.39$, $p{=}3.0{\times}10^{-5}$ \\
Zep (published)$^\S$ \citep{rasmussen2025zep} & $\sim$71\% & --- & --- & --- & --- \\
Supermemory \citep{supermemory2024} & 65.39\% & [62.98, 67.72] & 1{,}007 & \$0.0019 & $\chi^2=98.09$, $p{<}10^{-22}$ \\
Mem0 \citep{chhikara2025mem0} & 61.43\% & [58.97, 63.83] & 946 & \$0.0031 & $\chi^2=6.26$, $p{=}0.012$ \\
\bottomrule
\end{tabular}

\vspace{4pt}
{\footnotesize\raggedright
$^\ast$EverMemOS retrieval is served by proprietary cloud infrastructure (a 4-billion-parameter GPU-hosted embedder with a Neo4j graph store); the retrieval pipeline is not open source and cannot be run on local hardware. Reported numbers use pre-computed retrieval outputs provided by the EverMemOS team.\par
\vspace{2pt}
$^\dagger$Cost per correct reflects only the harness answer and judge API calls. EverMemOS retrieval compute runs on its own proprietary infrastructure and is not included; infrastructure requirements are discussed in \S\ref{sec:implementation}.\par
\vspace{2pt}
$^\ddagger$True Memory scores are 3-run means on the v0.6.0 pipeline. Individual run accuracies: Pro 92.79\%, 93.05\%, 93.05\%; Base 91.75\%, 92.08\%, 92.21\%; Edge 89.87\%, 89.55\%, 89.55\%. Wilson CIs and correct counts are computed from the mean correct count rounded to the nearest integer.\par
\vspace{2pt}
$^\S$Zep accuracy is from the published Zep paper \citep{rasmussen2025zep} using their own evaluation methodology, not our harness.\par}
\end{table*}

\begin{figure*}[htbp]
\centering
\begin{subfigure}[t]{0.48\textwidth}
\centering
\includegraphics[width=\textwidth]{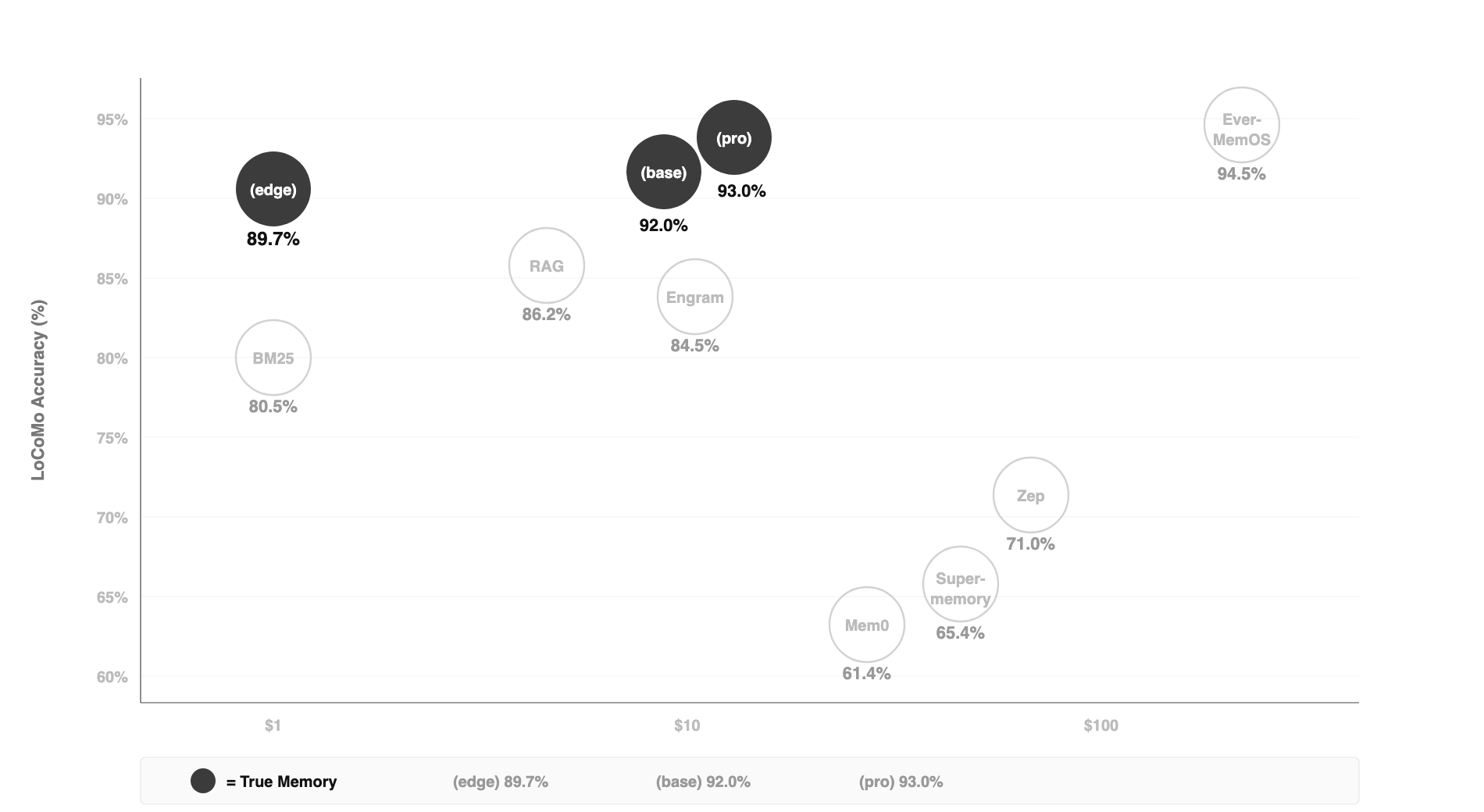}
\caption{\textbf{True Memory dominates the cost-accuracy Pareto frontier except for EverMemOS.} Horizontal axis: estimated monthly infrastructure cost (log scale). Vertical axis: LoCoMo accuracy.}
\label{fig:cost-pareto}
\end{subfigure}
\hfill
\begin{subfigure}[t]{0.48\textwidth}
\centering
\includegraphics[width=\textwidth]{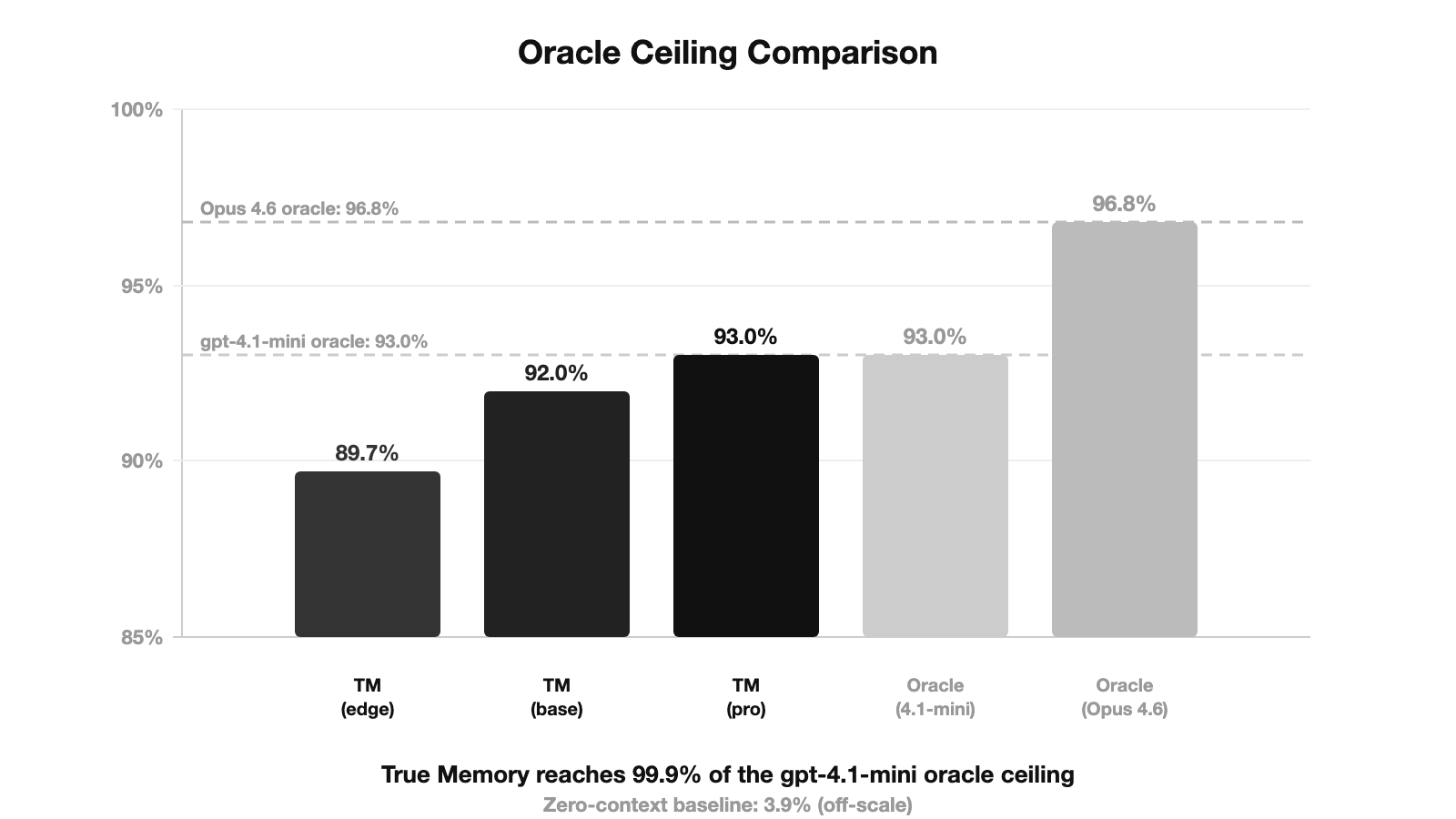}
\caption{\textbf{True Memory Pro reaches 99.9\% of the gpt-4.1-mini oracle ceiling.} Oracle-ceiling comparison on LoCoMo: True Memory Pro (v0.6.0, 3-run mean) reaches 99.9\% of the gpt-4.1-mini ceiling and 96.0\% of the Opus 4.6 ceiling.}
\label{fig:oracle-ceiling}
\end{subfigure}
\caption{Cost-accuracy and oracle-ceiling analysis on LoCoMo. \textbf{(a)} True Memory's three tiers cluster at the origin of the cost axis alongside EverMemOS and other open-source systems. Paid services (Mem0, Supermemory) incur monthly costs that shift them rightward. Every non-frontier system is strictly dominated. \textbf{(b)} True Memory Pro consumes 256-dimensional embeddings rather than the oracle conditions' 45.6 million input tokens per run.}

\vspace{1.5em}

\centering\footnotesize
\setlength{\tabcolsep}{3pt}
\captionof{table}{$2 \times 2$ contingency on the 357-question retrieval-bottleneck subset.}
\label{tab:contingency}
\begin{tabular}{@{}lrrr@{}}
\toprule
& Orac.\ corr. & Orac.\ incorr. & Total \\
\midrule
Retr.\ correct & 0 & 0 & 0 \\
Retr.\ incorrect & 330 & 27 & 357 \\
Column total & 330 & 27 & 357 \\
\bottomrule
\end{tabular}
\end{figure*}

\paragraph{Paired comparisons among the three True Memory configurations.} On v0.6.0, the 3-run mean gaps are: TM Pro vs.\ TM Base, $1.0$ pp; TM Base vs.\ TM Edge, $2.4$ pp; TM Pro vs.\ TM Edge, $3.4$ pp. The Pro--Base gap is small and may not be statistically resolved; the Base--Edge gap is consistent across all three runs. All three tiers cluster at the origin of the cost-accuracy Pareto frontier (Figure~\ref{fig:cost-pareto}).

\paragraph{Oracle ceiling.} The theoretical upper bound on LoCoMo is to supply the full conversation directly to the answer model, bypassing any memory system. Under \texttt{gpt-4.1-mini}, this full-context oracle reaches 92.99\% (1{,}432 of 1{,}540, 95\% CI $[91.60, 94.16]$) at a cost of 45.6 million input tokens per run, or \$0.0129 per correct answer, so True Memory Pro at 93.0\% reaches 99.9\% of oracle accuracy (Figure~\ref{fig:oracle-ceiling}) (unrounded: 92.96\% 3-run mean vs.\ 92.99\% oracle; both round to 93.0\%) at one-fifth the cost per correct. Under Claude Opus 4.6, the same full-context oracle reaches 96.75\% (1{,}490 of 1{,}540, CI $[95.75, 97.53]$), indicating that the architectural gain stacks with model gains rather than substituting for them. For reference, \texttt{gpt-4.1-mini} with no conversation context at all scores 3.90\% (60 of 1{,}540), which establishes the floor.

\subsection{Diagnostic: retrieval is the bottleneck}
\label{sec:diagnostic}

On an early iteration of True Memory, 357 LoCoMo questions were answered incorrectly. When the answer model was given the full conversation in place of retrieval results on those same 357 questions, 330 of 357 (92.44\%, 95\% Wilson CI $[89.22, 94.75]$) were then answered correctly. The $2 \times 2$ contingency of paired outcomes on this subset is reported in Table~\ref{tab:contingency}.

\paragraph{Per-category breakdown.} Grouping the 357 items by LoCoMo's four question categories and computing per-category recovery under the full-context oracle (95\% Wilson CIs shown):

\begin{center}\footnotesize
\setlength{\tabcolsep}{3pt}
\begin{tabular}{@{}lrrr@{}}
\toprule
Category & Recov. & Count & 95\% CI \\
\midrule
Cat 1 (single-sess.) & 94.0\% & 94/100 & [87.5, 97.2] \\
Cat 2 (multi-sess.)  & 87.8\% & 65/74  & [78.5, 93.5] \\
Cat 3 (knowl.-upd.)  & 86.2\% & 25/29  & [69.4, 94.5] \\
Cat 4 (temporal)     & 94.8\% & 146/154 & [90.1, 97.3] \\
\bottomrule
\end{tabular}
\end{center}

Category 3 (knowledge-update) is the smallest and the hardest-to-recover category, consistent with the intuition that contradiction-bearing questions are where a retrieval-only system is most likely to surface stale information even when the correcting utterance is present in storage.

\paragraph{Residual 8\% analysis.} 27 of 357 questions were still answered incorrectly under full context. A best-effort qualitative pass over the wrong rows in \texttt{full\_context\_gpt4mini.json} places roughly 20 of the 27 in the genuine-reasoning-failure bucket (arithmetic or multi-step inference over dates, counts, or causal chains), 5 in the temporal-annotation-ambiguity bucket (questions whose gold answer fixes a calendar date that the conversation only licenses approximately), and 2 in an ambiguous/unanswerable bucket; these counts are heuristic and should be treated as provisional.

\paragraph{Version note.} The 357-question diagnostic above was run on an early pipeline iteration. On the current v0.6.0 pipeline, True Memory Pro answers 1{,}432 of 1{,}540 correctly (3-run mean), leaving approximately 108 incorrect questions. The retrieval-bottleneck hypothesis is further supported: at 93.0\%, the system reaches 99.9\% of the \texttt{gpt-4.1-mini} oracle ceiling (92.99\%), so the remaining gap is dominated by questions the answer model itself cannot solve even with full context.

\paragraph{Horizon caveat.} This diagnostic works because a full LoCoMo conversation fits in current context windows. At longer horizons of weeks, months, or years of conversation, the full history cannot be loaded into any model's context window, so the diagnostic becomes unavailable and retrieval becomes the dominant scalable path. LoCoMo and similar benchmarks measure short-to-medium horizons, whereas True Memory's gated ingestion, consolidation, and accumulating speaker prior are designed to scale past where these benchmarks stop measuring.

\subsection{LongMemEval (cross-benchmark generalization)}
\label{sec:longmemeval}

Table~\ref{tab:longmemeval} reports systems on LongMemEval \citep{wu2024longmemeval}. True Memory Pro was evaluated on the v0.6.0 pipeline under \texttt{gpt-4.1-mini}; competitor rows are from a prior evaluation under the same harness conventions.

\begin{table*}[htbp]
\centering\small
\caption{LongMemEval accuracy with 95\% Wilson confidence intervals. $N = 500$. Evaluated on the LongMemEval strict variant (\texttt{longmemeval\_s.json}). True Memory Pro is a 3-run mean on the v0.6.0 pipeline; competitor rows are single-run.}
\label{tab:longmemeval}
\begin{tabular}{lrrrl}
\toprule
System & Accuracy & 95\% CI & Correct & Answer model \\
\midrule
\textbf{TM Pro} (oracle variant) & \textbf{92.0\%} & [89.43, 94.07] & 460 & \texttt{gpt-4.1-mini} \\
TM Pro (3-run mean, strict) & 87.8\% & [84.64, 90.38] & 439 & \texttt{gpt-4.1-mini} \\
RAG (ChromaDB)      & 87.0\% & [83.77, 89.67] & 435 & \texttt{gpt-4.1-mini} \\
EverMemOS$^\ast$ \citep{hu2026evermemos} & 83.0\% & [79.40, 86.10] & --- & \texttt{gpt-4o}$^\dagger$ \\
Engram & 82.2\% & [78.61, 85.30] & 411 & \texttt{gpt-4.1-mini} \\
BM25   & 81.6\% & [77.97, 84.75] & 408 & \texttt{gpt-4.1-mini} \\
Mem0 \citep{chhikara2025mem0}  & 66.0\% & [61.74, 70.02] & 330 & \texttt{gpt-4.1-mini} \\
\bottomrule
\end{tabular}

\vspace{4pt}
{\footnotesize\raggedright $^\ast$EverMemOS retrieval is served by proprietary cloud infrastructure and cannot be run on local hardware; reported number is from the EverMemOS paper.\par
\vspace{2pt}
$^\dagger$EverMemOS accuracy is from the published EverMemOS paper using \texttt{gpt-4o}, not our evaluation harness.\par}
\end{table*}

\paragraph{Per-category breakdown.} The multi-session category is the hardest for the retrieval pipeline, consistent with LoCoMo's per-category pattern; single-session categories score highest.

\paragraph{Cross-benchmark comparison.} True Memory Pro leads RAG-ChromaDB on LongMemEval by 0.8 pp (87.8\% vs.\ 87.0\%) and every other agent memory product by at least 4.8 pp (EverMemOS at 83.0\%). The result is consistent with the prior v0.2.0 evaluation (87.2\%), indicating that the retrieval pipeline upgrades in v0.6.0 maintain cross-benchmark accuracy while the primary gains appear on LoCoMo and BEAM-1M.

\subsection{BEAM-1M (long-horizon generalization)}
\label{sec:beam}

BEAM \citep{tavakoli2026beam} evaluates memory at conversation scales that exceed any model's context window. The 1M-token split contains 35 conversations with 20 questions each (700 total), spanning 10 memory-ability categories. Table~\ref{tab:beam} reports True Memory Pro (3-run mean) alongside the prior published result.

\begin{table}[htbp]
\centering\footnotesize
\caption{BEAM-1M accuracy (3-run mean). $N = 700$ (35 convs $\times$ 20 questions). Individual runs: 75.0\%, 78.1\%, 76.6\%.}
\label{tab:beam}
\begin{tabular}{@{}lrl@{}}
\toprule
System & Acc. & Answer model \\
\midrule
\textbf{TM Pro} (3-run) & \textbf{76.6\%} & gpt-4.1-mini \\
Hindsight \citep{latimer2025hindsight} & 73.9\% & proprietary \\
\bottomrule
\end{tabular}
\end{table}

\paragraph{Per-category breakdown.} Performance varies substantially across BEAM's ten memory-ability categories (3-run means): preference following 97.1\%, contradiction resolution 91.4\%, information extraction 91.4\%, summarization 89.5\%, instruction following 84.8\%, abstention 82.4\%, knowledge update 77.6\%, multi-session reasoning 67.1\%, temporal reasoning 64.8\%, event ordering 19.5\%. The top six categories exceed 80\%, while event ordering (19.5\%) is an outlier that requires chronological sequencing of events across sessions, a capability that the current pipeline does not specifically optimize for. Temporal reasoning (64.8\%) and multi-session reasoning (67.1\%) represent the next targets for improvement.

\paragraph{Significance.} BEAM-1M tests a qualitatively different regime from LoCoMo: at 1 million tokens per conversation, no model's context window holds the history, and the full-context oracle diagnostic of \S\ref{sec:diagnostic} is unavailable. True Memory's 76.6\% exceeds the prior published Hindsight result of 73.9\% by 2.7 pp, though we note that Hindsight uses a different answer model and the comparison is not controlled for that variable. The result demonstrates that the retrieval architecture generalizes from LoCoMo's short-to-medium horizons to the long-horizon regime the system was designed for.

\subsection{Abstention}
\label{sec:abstention}

Abstention is evaluated as one of the ten BEAM-1M categories (\S\ref{sec:beam}), where True Memory Pro scores 82.4\% (3-run mean).

\subsection{Comparison to EverMemOS}
\label{sec:evermemos}

EverMemOS outscores True Memory on LoCoMo at 94.48\%, 1.48 points above True Memory Pro's 3-run mean of 93.0\%. Both are retrieval-based architectures, and the gap at this margin is small. EverMemOS achieves the higher number by employing a 4-billion-parameter embedder served on GPUs, a Neo4j graph store, and a proprietary multi-service retrieval stack, whereas True Memory reaches 93.0\% with a single SQLite file on commodity CPU. The 1.5-point gap separates a GPU-served, graph-backed retrieval pipeline from one that runs on commodity CPU.

\section{Ablations}
\label{sec:ablations}

This section reports a 56-configuration grid study on LoCoMo that evaluates the sensitivity of True Memory's accuracy to component choice within the retrieval pipeline.

\paragraph{Grid design.} We evaluated 56 combinations of 7 embedder classes $\times$ 8 reranker options (including a no-reranker control) on LoCoMo, using the evaluation harness of \S\ref{sec:results}, with hypothetical-document query expansion enabled in all 56 runs. Embedder classes spanned static 256-dimensional embeddings, dense transformer embeddings at 256, 512, 768, and 1024 dimensions, and a published retrieval-tuned dense embedding, while reranker options spanned lightweight cross-encoders, modernbert-based cross-encoders, bge-family cross-encoders, and a no-reranker control.

\paragraph{Aggregate results.} Accuracy across the full grid ranged from 89.9\% at the worst configuration to 93.1\% at the best, a 3.2-percentage-point spread, with a mean of 91.26\% and a median of 91.20\%. Fifty-three of the 56 configurations achieved 90.0\% or higher.

\begin{table*}[htbp]
\centering\small
\caption{Grid-wide and subfamily statistics on LoCoMo (56 configurations, 1{,}540 questions).$^\star$}
\label{tab:ablation-agg}
\begin{tabular}{lrrrrrr}
\toprule
Scope & Cells & Best & Worst & Mean & Median & Spread \\
\midrule
Full 56-configuration grid & 56 & 93.1\% & 89.9\% & 91.26\% & 91.20\% & 3.2 pp \\
Matryoshka-256d $\times$ reranker subfamily & 8 & 92.3\% & 91.0\% & 91.46\% & 91.4\% & \textbf{1.3 pp} \\
\bottomrule
\end{tabular}
\end{table*}

\begin{figure*}[htbp]
\centering
\includegraphics[width=0.9\textwidth]{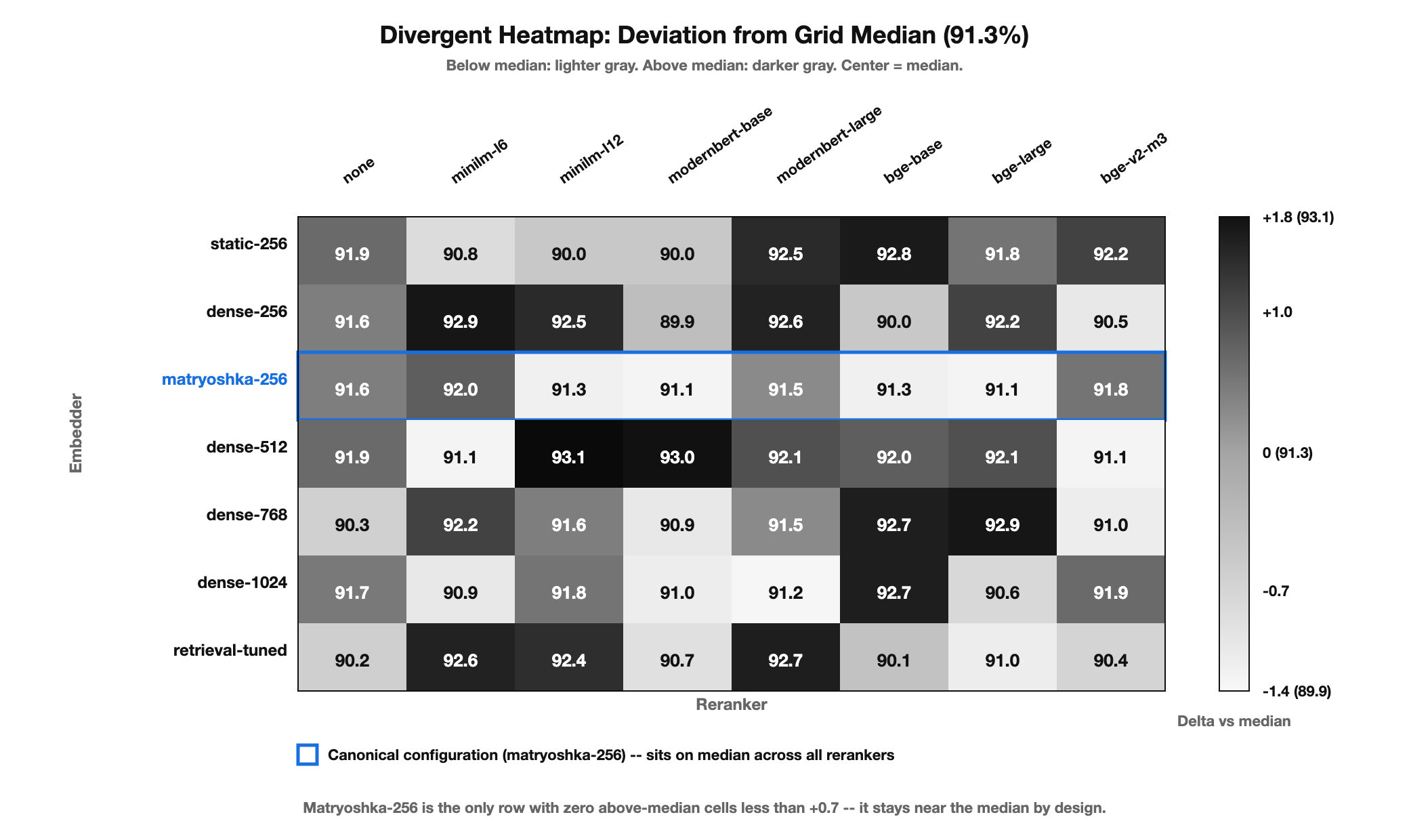}
\caption{\textbf{The retrieval pipeline moves accuracy by at most 3.2 pp across 56 configurations; within the Matryoshka subfamily, at most 1.3 pp.}$^\star$ Heatmap of the 56-configuration LoCoMo grid. Rows: 7 embedder classes. Columns: 8 reranker options including a no-reranker control. Cell color encodes accuracy from 89.9\% (grid worst) to 93.1\% (grid best). The Matryoshka-trained 256-dimensional embedder row shows the 1.3-percentage-point subfamily spread referenced in the abstract.}
\label{fig:ablation-heatmap}
\end{figure*}

{\footnotesize $^\star$Ablation data collected on the v0.4.0 pipeline. Absolute scores may differ on v0.6.0; relative rankings of embedder/reranker combinations are expected to hold (see footnote in Query Expansion Ablation below).\par}

\paragraph{Subfamily behavior.} Within the subfamily of a Matryoshka-trained 256-dimensional embedding \citep{kusupati2022matryoshka} paired with any reranker, the total accuracy spread is 1.3 percentage points (best \texttt{qwen3\_256d $\times$ zerank1} at 92.3\%, worst \texttt{qwen3\_256d $\times$ bge\_large} and \texttt{qwen3\_256d $\times$ mxbai\_large} at 91.0\%), so that changing the reranker identity within this family moves accuracy by at most 1.3 pp. This is the narrow-range figure referenced in the abstract.

\enlargethispage{\baselineskip}
\paragraph{Query expansion ablation.} We additionally ran the Base configuration with and without hypothetical-document query expansion \citep{gao2023hyde}; enabling query expansion adds 1.0 pp on LoCoMo (92.0\% without, 93.0\% with, v0.6.0 3-run means).\footnotemark

\paragraph{Interpretation.}\footnotetext{Ablation data in Table~\ref{tab:ablation-agg}, Table~\ref{tab:ablation-full}, and Figure~\ref{fig:ablation-heatmap} were collected on the v0.4.0 pipeline; absolute scores may differ on v0.6.0, but relative rankings of embedder/reranker combinations are expected to hold as these choices are independent of the gate, L0, and dedup changes introduced in v0.5.0 and v0.6.0.} The grid provides a falsifiable architectural claim: if retrieval pipeline quality were primarily a matter of choosing the right embedder or reranker, we would expect wide variance across components, which is not what we observe. Within the authoritative configuration family (Matryoshka-trained 256-dimensional embedding plus any reranker), 1.3 percentage points separates the best from the worst, and within the full 56-configuration grid, 3.2 points separates the best from the worst; both ranges sit entirely above every prior published agent memory system measured in \S\ref{sec:results}. What distinguishes True Memory from extraction-based competitors is not which embedder or reranker is chosen, but whether the retrieval pipeline exists as an architectural center at all.

\section{Discussion}
\label{sec:discussion}

The claim this paper advances is that retrieval, rather than storage, is what determines the behavior of a memory system. The evidence is roughly 30 percentage points of LoCoMo accuracy separating True Memory from every extraction-based competitor, and roughly 19 percentage points separating its lightest configuration (Edge, 89.7\%) from every extraction-based competitor. Accuracy variance across 56 component combinations within the retrieval-based regime is 3.2 percentage points, and within the top-performing subfamily it is 1.3, so that architecture separates from its absence by roughly an order of magnitude more than component choice separates from itself.

The three retrieval baselines included in \S\ref{sec:results}, namely BM25 (80.5\%), Engram (84.5\%), and RAG-ChromaDB (86.2\%), cluster within a 5.7-pp band on LoCoMo, forming a ceiling for systems that retrieve stored content without reasoning about it at query time. All three preserve events verbatim and retrieve by lexical or semantic similarity alone; none filters noise at ingestion, detects contradictions between temporally separated messages, resolves relative time references, or conditions retrieval on speaker identity. The extraction-based systems (Mem0 at 61.4\%, Supermemory at 65.4\%) fall roughly 20 percentage points below this ceiling because they discard the verbatim substrate entirely. True Memory Pro exceeds every baseline by 6.8 to 12.5 percentage points depending on the comparison, and even the lightweight Edge configuration exceeds RAG by 3.5 pp. The ablation of \S\ref{sec:ablations} shows that the gap is not attributable to any single component: within the top-performing embedder subfamily, swapping the reranker moves accuracy by at most 1.3 pp, yet every configuration in that family exceeds the best baseline by at least 4.8 pp. Three tiers of system emerge: extraction-based systems that lose information at ingestion, retrieval baselines that preserve it but surface it without query-time reasoning, and a retrieval-centered pipeline that preserves information and applies multi-stage ranking at recall. The gap between tiers is consistently wider than the gap between components within any single tier, so that the choice of retrieval architecture dominates the choice of any individual component within it.

Two open questions define the most immediate directions for future work: the encoding gate, and horizon.

\paragraph{Encoding gate.} The gate's three-signal combination achieves AUC 0.816 on a held-out evaluation set (validated in a 200-variant sweep; \texttt{truememory/ingest/encoding\_gate.py}), but current benchmarks inject fixed conversation transcripts and reward total recall, and therefore cannot score a system that chooses what to ingest. True Memory runs in production with selective ingestion active, and all reported results disable it for comparability; whether the gated system beats the ungated one on accuracy, and by how much, is an empirical question no public benchmark currently answers. We regard this as a gap in the evaluation literature rather than a limitation of the architecture.

\paragraph{Horizon.} LoCoMo tests memory at scales that fit in a frontier model's context window, which is why a full-context oracle remains computable. BEAM-1M (\S\ref{sec:beam}) begins to probe the long-horizon regime: at 1 million tokens per conversation, no model's context window holds the history, and True Memory Pro reaches 76.6\% (3-run mean). Preliminary single-run results on BEAM-10M (10 conversations at 10 million tokens each, approximately 20{,}000 messages per conversation, $N = 200$) reach 65.0\%, an 11.6-pp drop from the 1M split. The degradation is not uniform: knowledge-update accuracy improves from 77.6\% at 1M to 90.0\% at 10M, suggesting that the retrieval pipeline's contradiction-resolution mechanism benefits from deeper conversation history, while temporal reasoning (30.0\%) and event ordering (5.0\%) degrade sharply, consistent with the same weakness observed at 1M. A full 3-run evaluation on BEAM-10M is in progress. The agent-memory setting, however, extends to weeks, months, and years of conversation, scales at which no oracle is measurable and no benchmark yet exists. The architecture of \S\ref{sec:architecture} is engineered for that regime.

Verbatim event preservation yields a further architectural consequence: any scoring function introduced after ingestion can be evaluated retroactively against the full event history. A salience weighting defined later for a new domain, a retention classification, or any relevance function that did not exist at ingestion all operate on the same substrate the retrieval pipeline uses. Systems that commit to a scoring function at ingestion lack this property.

\section{Conclusion}
\label{sec:conclusion}

In this work we presented \textbf{True Memory}, an agent memory architecture that treats retrieval, rather than storage, as the architectural center. The system is organized as six cooperating layers over a substrate of events preserved verbatim, with an encoding gate at ingestion whose three signals (novelty, salience, prediction error) are computed through independent mechanisms, two of which query the same stored message substrate that serves retrieval.

On LoCoMo under a matched \texttt{gpt-4.1-mini} answer model, True Memory Pro reaches 93.0\% (3-run mean), trailing only the cloud-served EverMemOS by 1.5 points while running as a single SQLite file on commodity CPU. On LongMemEval, the system reaches 87.8\%, leading every agent memory product by at least 4.8 pp. On BEAM-1M, at 1-million-token conversation scale, True Memory Pro reaches 76.6\%, above the prior published Hindsight result of 73.9\%. A 56-configuration ablation shows that component choice within the retrieval pipeline moves accuracy by an order of magnitude less than the retrieval architecture itself moves it relative to extraction-based baselines, and a retrieval-bottleneck diagnostic locates the remaining errors in retrieval rather than storage.

Two directions follow. First, evaluation instruments for selective ingestion: the encoding gate (AUC 0.816, validated in a 200-variant sweep) is disabled throughout this paper because no public benchmark scores a system that chooses what to ingest, and closing this gap requires new benchmarks rather than new architectures. Second, deeper long-horizon evaluation: BEAM-1M begins to probe the regime the architecture was designed for, but weeks, months, and years of conversation remain beyond any current benchmark's reach.

The code, evaluation harness, and benchmark outputs are available at {\small\url{https://github.com/buildingjoshbetter/TrueMemory}}.

\section*{Limitations}

This work has three principal limitations. First, the encoding gate is disabled throughout all benchmark evaluations because no existing benchmark scores selective ingestion; the gate's contribution to end-to-end accuracy therefore remains unmeasured. Second, the longest evaluation corpus (BEAM-1M) spans approximately one million tokens, whereas the architecture is designed for conversation histories that grow over weeks or months; no public benchmark probes this regime. Third, all reported scores use a semantic-match judge, which is more lenient than exact-match scoring; rankings across systems are valid but absolute accuracy numbers should not be compared directly to strict-match baselines without adjustment.

\bibliographystyle{plainnat}

\begin{thebibliography}{32}
\providecommand{\natexlab}[1]{#1}
\providecommand{\url}[1]{\texttt{#1}}
\expandafter\ifx\csname urlstyle\endcsname\relax
  \providecommand{\doi}[1]{doi: #1}\else
  \providecommand{\doi}{doi: \begingroup \urlstyle{rm}\Url}\fi

\bibitem[Bartlett(1932)]{bartlett1932remembering}
Frederic~Charles Bartlett.
\newblock \emph{Remembering: {A} Study in Experimental and Social Psychology}.
\newblock Cambridge University Press, 1932.

\bibitem[Cahill and McGaugh(1995)]{cahill1995novel}
Larry Cahill and James~L. McGaugh.
\newblock A novel demonstration of enhanced memory associated with emotional
  arousal.
\newblock \emph{Consciousness and Cognition}, 4\penalty0 (4):\penalty0
  410--421, 1995.

\bibitem[Chhikara et~al.(2025)Chhikara, Khant, Aryan, Singh, and
  Yadav]{chhikara2025mem0}
Prateek Chhikara, Dev Khant, Saket Aryan, Taranjeet Singh, and Deshraj Yadav.
\newblock {Mem0}: Building production-ready {AI} agents with scalable long-term
  memory.
\newblock \emph{arXiv preprint arXiv:2504.19413}, 2025.

\bibitem[Cormack et~al.(2009)Cormack, Clarke, and B{\"u}ttcher]{cormack2009rrf}
Gordon~V. Cormack, Charles L.~A. Clarke, and Stefan B{\"u}ttcher.
\newblock Reciprocal rank fusion outperforms {Condorcet} and individual rank
  learning methods.
\newblock In \emph{Proceedings of the 32nd International ACM SIGIR Conference
  on Research and Development in Information Retrieval}, pages 758--759, 2009.

\bibitem[Craik and Lockhart(1972)]{craik1972levels}
Fergus I.~M. Craik and Robert~S. Lockhart.
\newblock Levels of processing: {A} framework for memory research.
\newblock \emph{Journal of Verbal Learning and Verbal Behavior}, 11\penalty0
  (6):\penalty0 671--684, 1972.

\bibitem[Gallagher and Frith(2003)]{gallagher2003theory}
Helen~L. Gallagher and Christopher~D. Frith.
\newblock Functional imaging of ``theory of mind''.
\newblock \emph{Trends in Cognitive Sciences}, 7\penalty0 (2):\penalty0 77--83,
  2003.

\bibitem[Gao et~al.(2023)Gao, Ma, Lin, and Callan]{gao2023hyde}
Luyu Gao, Xueguang Ma, Jimmy Lin, and Jamie Callan.
\newblock Precise zero-shot dense retrieval without relevance labels.
\newblock In \emph{Proceedings of the 61st Annual Meeting of the Association
  for Computational Linguistics}, pages 1762--1777, 2023.

\bibitem[Garcia(2024)]{garcia2024sqlitevec}
Alex Garcia.
\newblock sqlite-vec: {A} vector search {SQLite} extension.
\newblock \url{https://github.com/asg017/sqlite-vec}, 2024.
\newblock Accessed 2026-04-15.

\bibitem[Hipp and contributors(2000)]{hipp2000sqlite}
D.~Richard Hipp and contributors.
\newblock {SQLite} and the {FTS5} full-text-search module.
\newblock \url{https://www.sqlite.org/fts5.html}, 2000.
\newblock SQLite released in 2000; the FTS5 module was added in 2015. Accessed
  2026-04-15.

\bibitem[Hu et~al.(2026)Hu, Gao, Zhou, Xu, Bai, Li, Zhang, Li, Zhang, Bing, and
  Deng]{hu2026evermemos}
C.~Hu, X.~Gao, Z.~Zhou, D.~Xu, Y.~Bai, X.~Li, H.~Zhang, T.~Li, C.~Zhang,
  L.~Bing, and Y.~Deng.
\newblock {EverMemOS}: {A} self-organizing memory operating system for
  structured long-horizon reasoning, 2026.

\bibitem[Kusupati et~al.(2022)Kusupati, Bhatt, Rege, Wallingford, Sinha,
  Ramanujan, Howard-Snyder, Chen, Kakade, Jain, and
  Farhadi]{kusupati2022matryoshka}
Aditya Kusupati, Gantavya Bhatt, Aniket Rege, Matthew Wallingford, Aditya
  Sinha, Vivek Ramanujan, William Howard-Snyder, Kaifeng Chen, Sham Kakade,
  Prateek Jain, and Ali Farhadi.
\newblock Matryoshka representation learning.
\newblock In \emph{Advances in Neural Information Processing Systems 35}, pages
  30233--30249, 2022.

\bibitem[Latimer et~al.(2025)Latimer, Boschi, Neeser, Bartholomew, Srivastava,
  Wang, and Ramakrishnan]{latimer2025hindsight}
Chris Latimer, Nicol{\'o} Boschi, Andrew Neeser, Chris Bartholomew, Gaurav
  Srivastava, Xuan Wang, and Naren Ramakrishnan.
\newblock Hindsight is 20/20: Building agent memory that retains, recalls, and
  reflects.
\newblock \emph{arXiv preprint arXiv:2512.12818}, 2025.

\bibitem[Lewis et~al.(2020)Lewis, Perez, Piktus, Petroni, Karpukhin, Goyal,
  K{\"u}ttler, Lewis, Yih, Rockt{\"a}schel, Riedel, and Kiela]{lewis2020rag}
Patrick Lewis, Ethan Perez, Aleksandra Piktus, Fabio Petroni, Vladimir
  Karpukhin, Naman Goyal, Heinrich K{\"u}ttler, Mike Lewis, Wen-tau Yih, Tim
  Rockt{\"a}schel, Sebastian Riedel, and Douwe Kiela.
\newblock Retrieval-augmented generation for knowledge-intensive {NLP} tasks.
\newblock In \emph{Advances in Neural Information Processing Systems 33}, pages
  9459--9474, 2020.

\bibitem[Li et~al.(2026)Li, Li, Yu, Ding, Lin, Wang, and Zhou]{li2026query}
Yuqing Li, Jiangnan Li, Mo~Yu, Guoxuan Ding, Zheng Lin, Weiping Wang, and Jie Zhou.
\newblock Query-focused and memory-aware reranker for long context processing.
\newblock \emph{arXiv preprint arXiv:2602.12192}, 2026.

\bibitem[Maharana et~al.(2024)Maharana, Lee, Tulyakov, Bansal, Barbieri, and
  Fang]{maharana2024}
Adyasha Maharana, Dong-Ho Lee, Sergey Tulyakov, Mohit Bansal, Francesco
  Barbieri, and Yuwei Fang.
\newblock Evaluating very long-term conversational memory of {LLM} agents.
\newblock In \emph{Proceedings of the 62nd Annual Meeting of the Association
  for Computational Linguistics}, pages 13851--13870, 2024.
\newblock arXiv:2402.17753.

\bibitem[Malkov and Yashunin(2018)]{malkov2018hnsw}
Yu.~A. Malkov and D.~A. Yashunin.
\newblock Efficient and robust approximate nearest neighbor search using
  hierarchical navigable small world graphs.
\newblock \emph{IEEE Transactions on Pattern Analysis and Machine
  Intelligence}, 42\penalty0 (4):\penalty0 824--836, 2018.
\newblock arXiv:1603.09320.

\bibitem[McClelland et~al.(1995)McClelland, McNaughton, and
  O'Reilly]{mcclelland1995cls}
James~L. McClelland, Bruce~L. McNaughton, and Randall~C. O'Reilly.
\newblock Why there are complementary learning systems in the hippocampus and
  neocortex: Insights from the successes and failures of connectionist models
  of learning and memory.
\newblock \emph{Psychological Review}, 102\penalty0 (3):\penalty0 419--457,
  1995.

\bibitem[Nogueira and Cho(2019)]{nogueira2019bertrerank}
Rodrigo Nogueira and Kyunghyun Cho.
\newblock Passage re-ranking with {BERT}.
\newblock \emph{arXiv preprint arXiv:1901.04085}, 2019.

\bibitem[Packer et~al.(2023)Packer, Wooders, Lin, Fang, Patil, Stoica, and
  Gonzalez]{packer2023memgpt}
Charles Packer, Sarah Wooders, Kevin Lin, Vivian Fang, Shishir~G. Patil, Ion
  Stoica, and Joseph~E. Gonzalez.
\newblock {MemGPT}: Towards {LLMs} as operating systems.
\newblock \emph{arXiv preprint arXiv:2310.08560}, 2023.

\bibitem[{Pinecone} et~al.(2026){Pinecone}, {Chroma}, {Weaviate}, and
  {Qdrant}]{vectordb2026}
{Pinecone}, {Chroma}, {Weaviate}, and {Qdrant}.
\newblock Vector database products built on {HNSW}-class indexing, 2026.
\newblock Representative commercial and open-source vector database products;
  product documentation at \url{https://www.pinecone.io},
  \url{https://www.trychroma.com}, \url{https://weaviate.io}, and
  \url{https://qdrant.tech}, accessed 2026-04-15.

\bibitem[{Qwen Team}(2025)]{qwen2025embedding}
{Qwen Team}.
\newblock {Qwen3-Embedding}: Advanced text embedding and reranking through
  foundation models, 2025.

\bibitem[Rao and Ballard(1999)]{rao1999predictive}
Rajesh P.~N. Rao and Dana~H. Ballard.
\newblock Predictive coding in the visual cortex: {A} functional interpretation
  of some extra-classical receptive-field effects.
\newblock \emph{Nature Neuroscience}, 2\penalty0 (1):\penalty0 79--87, 1999.

\bibitem[Rasmussen et~al.(2025)Rasmussen, Paliychuk, Beauvais, Ryan, and
  Chalef]{rasmussen2025zep}
Preston Rasmussen, Pavlo Paliychuk, Travis Beauvais, Jack Ryan, and Daniel
  Chalef.
\newblock {Zep}: {A} temporal knowledge graph architecture for agent memory,
  2025.

\bibitem[Schacter(2001)]{schacter2001seven}
Daniel~L. Schacter.
\newblock \emph{The Seven Sins of Memory: {H}ow the Mind Forgets and
  Remembers}.
\newblock Houghton Mifflin, 2001.

\bibitem[Squire and Alvarez(1995)]{squire1995retrograde}
Larry~R. Squire and Pablo Alvarez.
\newblock Retrograde amnesia and memory consolidation: {A} neurobiological
  perspective.
\newblock \emph{Current Opinion in Neurobiology}, 5\penalty0 (2):\penalty0
  169--177, 1995.

\bibitem[Sukhbaatar et~al.(2015)Sukhbaatar, Szlam, Weston, and
  Fergus]{sukhbaatar2015memory}
Sainbayar Sukhbaatar, Arthur Szlam, Jason Weston, and Rob Fergus.
\newblock End-to-end memory networks.
\newblock In \emph{Advances in Neural Information Processing Systems 28}, pages
  2440--2448, 2015.
\newblock arXiv:1503.08895.

\bibitem[{Supermemory}(2024)]{supermemory2024}
{Supermemory}.
\newblock Supermemory.
\newblock \url{https://supermemory.ai/}, 2024.
\newblock Commercial agent-memory service. Documentation at
  \url{https://docs.supermemory.ai}. Accessed 2026-04-15.

\bibitem[Tavakoli et~al.(2026)Tavakoli, Salemi, Ye, Abdalla, Zamani, and
  Mitchell]{tavakoli2026beam}
Mohammad Tavakoli, Alireza Salemi, Carrie Ye, Mohamed Abdalla, Hamed Zamani,
  and J.~Ross Mitchell.
\newblock Beyond a million tokens: Benchmarking and enhancing long-term memory
  in {LLMs}.
\newblock In \emph{International Conference on Learning Representations
  (ICLR)}, 2026.
\newblock arXiv:2510.27246.

\bibitem[Tulkens and van Dongen(2024)]{tulkens2024model2vec}
St{\'e}phan Tulkens and Thomas van Dongen.
\newblock Model2vec: {F}ast static embeddings from sentence transformers.
\newblock \url{https://github.com/MinishLab/model2vec}, 2024.
\newblock Accessed 2026-04-15.

\bibitem[Tulving(1972)]{tulving1972episodic}
Endel Tulving.
\newblock Episodic and semantic memory.
\newblock In Endel Tulving and Wayne Donaldson, editors, \emph{Organization of
  Memory}, pages 381--403. Academic Press, 1972.

\bibitem[Wu et~al.(2025)Wu, Wang, Yu, Zhang, Chang, and Yu]{wu2024longmemeval}
Di~Wu, Hongwei Wang, Wenhao Yu, Yichong Zhang, Kai-Wei Chang, and Dong Yu.
\newblock {LongMemEval}: Benchmarking chat assistants on long-term interactive
  memory.
\newblock In \emph{International Conference on Learning Representations
  (ICLR)}, 2025.
\newblock arXiv:2410.10813.

\bibitem[{Zep AI}(2024)]{zepai2024graphiti}
{Zep AI}.
\newblock Graphiti: {A} temporal knowledge graph framework for ai agents.
\newblock \url{https://github.com/getzep/graphiti}, 2024.
\newblock Accessed 2026-04-15.

\end{thebibliography}

\appendix
\onecolumn
\section{Complete 56-configuration ablation data}
\label{app:ablation-full}

Table~\ref{tab:ablation-full} lists every cell of the 7 embedder $\times$ 8 reranker grid referenced in \S\ref{sec:ablations}, sorted by LoCoMo accuracy. The aggregate statistics in Table~\ref{tab:ablation-agg} and the heatmap in Figure~\ref{fig:ablation-heatmap} are computed directly from these 56 rows.

\vspace{1em}
\begin{center}
\scriptsize
\renewcommand{\arraystretch}{0.82}
\setlength{\tabcolsep}{3pt}
\captionof{table}{Per-configuration results across the 56-cell LoCoMo grid.$^\star$ Wilson 95\% CIs computed from correct-count over $N = 1{,}540$. Retrieval latency is the harness-reported average per query.}
\label{tab:ablation-full}
\begin{tabular}{llrlrrlr}
\toprule
Config & Embedder & Dim & Reranker & Accuracy & Correct & 95\% Wilson CI & Retrieval \\
\midrule
C01 & \texttt{zembed1} & --- & \texttt{mxbai\_large} & 93.1\% & 1433/1540 & [91.67, 94.22] & 3.29\,s \\
C02 & \texttt{zembed1} & --- & \texttt{qwen3\_reranker} & 92.9\% & 1430/1540 & [91.46, 94.04] & 5.34\,s \\
C03 & \texttt{zembed1} & --- & \texttt{gte\_reranker} & 92.6\% & 1426/1540 & [91.18, 93.80] & 2.69\,s \\
C04 & \texttt{qwen3\_256d} & 256 & \texttt{zerank1} & 92.3\% & 1421/1540 & [90.83, 93.50] & 3.11\,s \\
C05 & \texttt{zembed1} & --- & \texttt{bge\_v2\_m3} & 92.2\% & 1420/1540 & [90.76, 93.44] & 4.50\,s \\
C06 & \texttt{nomic\_768d} & 768 & \texttt{zerank1} & 92.0\% & 1417/1540 & [90.55, 93.26] & 2.53\,s \\
C07 & \texttt{qwen3\_512d} & 512 & \texttt{zerank1} & 92.0\% & 1417/1540 & [90.55, 93.26] & 3.06\,s \\
C08 & \texttt{gte\_modernbert\_768d} & 768 & \texttt{gte\_reranker} & 91.9\% & 1415/1540 & [90.41, 93.15] & 2.58\,s \\
C09 & \texttt{qwen3\_512d} & 512 & \texttt{qwen3\_reranker} & 91.9\% & 1415/1540 & [90.41, 93.15] & 4.40\,s \\
C10 & \texttt{zembed1} & --- & \texttt{zerank1} & 91.9\% & 1416/1540 & [90.48, 93.20] & 4.08\,s \\
C11 & \texttt{gte\_modernbert\_768d} & 768 & \texttt{bge\_v2\_m3} & 91.8\% & 1414/1540 & [90.34, 93.09] & 3.42\,s \\
C12 & \texttt{qwen3\_256d} & 256 & \texttt{gte\_reranker} & 91.8\% & 1414/1540 & [90.34, 93.09] & 2.46\,s \\
C13 & \texttt{gte\_modernbert\_768d} & 768 & \texttt{qwen3\_reranker} & 91.7\% & 1412/1540 & [90.20, 92.97] & 4.13\,s \\
C14 & \texttt{qwen3\_1024d} & 1024 & \texttt{qwen3\_reranker} & 91.7\% & 1412/1540 & [90.20, 92.97] & 4.26\,s \\
C15 & \texttt{qwen3\_256d} & 256 & \texttt{miniml12} & 91.7\% & 1412/1540 & [90.20, 92.97] & 1.92\,s \\
C16 & \texttt{gte\_modernbert\_768d} & 768 & \texttt{zerank1} & 91.6\% & 1410/1540 & [90.06, 92.85] & 2.73\,s \\
C17 & \texttt{qwen3\_512d} & 512 & \texttt{gte\_reranker} & 91.6\% & 1410/1540 & [90.06, 92.85] & 2.73\,s \\
C18 & \texttt{zembed1} & --- & \texttt{bge\_large} & 91.6\% & 1410/1540 & [90.06, 92.85] & 4.42\,s \\
C19 & \texttt{qwen3\_512d} & 512 & \texttt{bge\_v2\_m3} & 91.5\% & 1409/1540 & [89.99, 92.79] & 3.34\,s \\
C20 & \texttt{qwen3\_512d} & 512 & \texttt{mxbai\_large} & 91.5\% & 1409/1540 & [89.99, 92.79] & 2.34\,s \\
C21 & \texttt{model2vec\_256d} & 256 & \texttt{miniml12} & 91.4\% & 1407/1540 & [89.86, 92.67] & 9.41\,s \\
C22 & \texttt{qwen3\_256d} & 256 & \texttt{qwen3\_reranker} & 91.4\% & 1407/1540 & [89.86, 92.67] & 4.36\,s \\
C23 & \texttt{qwen3\_512d} & 512 & \texttt{bge\_large} & 91.4\% & 1408/1540 & [89.93, 92.73] & 3.34\,s \\
C24 & \texttt{model2vec\_256d} & 256 & \texttt{bge\_large} & 91.3\% & 1406/1540 & [89.79, 92.61] & 3.39\,s \\
C25 & \texttt{nomic\_768d} & 768 & \texttt{qwen3\_reranker} & 91.3\% & 1406/1540 & [89.79, 92.61] & 4.01\,s \\
C26 & \texttt{qwen3\_256d} & 256 & \texttt{bge\_v2\_m3} & 91.3\% & 1406/1540 & [89.79, 92.61] & 3.50\,s \\
C27 & \texttt{qwen3\_512d} & 512 & \texttt{miniml12} & 91.3\% & 1406/1540 & [89.79, 92.61] & 2.11\,s \\
C28 & \texttt{gte\_modernbert\_768d} & 768 & \texttt{bge\_large} & 91.2\% & 1405/1540 & [89.72, 92.55] & 3.14\,s \\
C29 & \texttt{gte\_modernbert\_768d} & 768 & \texttt{miniml12} & 91.2\% & 1404/1540 & [89.65, 92.49] & 1.86\,s \\
C30 & \texttt{gte\_modernbert\_768d} & 768 & \texttt{no\_reranker} & 91.2\% & 1404/1540 & [89.65, 92.49] & 1.71\,s \\
C31 & \texttt{nomic\_768d} & 768 & \texttt{bge\_large} & 91.2\% & 1404/1540 & [89.65, 92.49] & 3.56\,s \\
C32 & \texttt{qwen3\_256d} & 256 & \texttt{no\_reranker} & 91.2\% & 1404/1540 & [89.65, 92.49] & 1.81\,s \\
C33 & \texttt{gte\_modernbert\_768d} & 768 & \texttt{mxbai\_large} & 91.1\% & 1403/1540 & [89.58, 92.43] & 2.63\,s \\
C34 & \texttt{qwen3\_512d} & 512 & \texttt{no\_reranker} & 91.1\% & 1403/1540 & [89.58, 92.43] & 1.89\,s \\
C35 & \texttt{model2vec\_256d} & 256 & \texttt{zerank1} & 91.0\% & 1402/1540 & [89.51, 92.37] & 3.78\,s \\
C36 & \texttt{nomic\_768d} & 768 & \texttt{no\_reranker} & 91.0\% & 1402/1540 & [89.51, 92.37] & 1.74\,s \\
C37 & \texttt{qwen3\_256d} & 256 & \texttt{bge\_large} & 91.0\% & 1401/1540 & [89.44, 92.31] & 3.48\,s \\
C38 & \texttt{qwen3\_256d} & 256 & \texttt{mxbai\_large} & 91.0\% & 1401/1540 & [89.44, 92.31] & 2.30\,s \\
C39 & \texttt{nomic\_768d} & 768 & \texttt{bge\_v2\_m3} & 90.9\% & 1400/1540 & [89.37, 92.24] & 3.04\,s \\
C40 & \texttt{nomic\_768d} & 768 & \texttt{miniml12} & 90.9\% & 1400/1540 & [89.37, 92.24] & 1.92\,s \\
C41 & \texttt{qwen3\_1024d} & 1024 & \texttt{bge\_v2\_m3} & 90.9\% & 1400/1540 & [89.37, 92.24] & 3.63\,s \\
C42 & \texttt{qwen3\_1024d} & 1024 & \texttt{miniml12} & 90.9\% & 1400/1540 & [89.37, 92.24] & 2.09\,s \\
C43 & \texttt{nomic\_768d} & 768 & \texttt{mxbai\_large} & 90.8\% & 1398/1540 & [89.23, 92.12] & 2.34\,s \\
C44 & \texttt{qwen3\_1024d} & 1024 & \texttt{no\_reranker} & 90.8\% & 1399/1540 & [89.30, 92.18] & 1.86\,s \\
C45 & \texttt{qwen3\_1024d} & 1024 & \texttt{zerank1} & 90.8\% & 1399/1540 & [89.30, 92.18] & 3.62\,s \\
C46 & \texttt{qwen3\_1024d} & 1024 & \texttt{gte\_reranker} & 90.7\% & 1397/1540 & [89.16, 92.06] & 2.62\,s \\
C47 & \texttt{qwen3\_1024d} & 1024 & \texttt{mxbai\_large} & 90.7\% & 1397/1540 & [89.16, 92.06] & 2.25\,s \\
C48 & \texttt{model2vec\_256d} & 256 & \texttt{gte\_reranker} & 90.6\% & 1395/1540 & [89.02, 91.94] & 2.36\,s \\
C49 & \texttt{zembed1} & --- & \texttt{miniml12} & 90.6\% & 1395/1540 & [89.02, 91.94] & 26.86\,s \\
C50 & \texttt{model2vec\_256d} & 256 & \texttt{qwen3\_reranker} & 90.5\% & 1394/1540 & [88.95, 91.88] & 4.21\,s \\
C51 & \texttt{zembed1} & --- & \texttt{no\_reranker} & 90.5\% & 1393/1540 & [88.88, 91.82] & 1.54\,s \\
C52 & \texttt{nomic\_768d} & 768 & \texttt{gte\_reranker} & 90.4\% & 1392/1540 & [88.82, 91.76] & 2.54\,s \\
C53 & \texttt{qwen3\_1024d} & 1024 & \texttt{bge\_large} & 90.1\% & 1387/1540 & [88.47, 91.46] & 3.38\,s \\
C54 & \texttt{model2vec\_256d} & 256 & \texttt{bge\_v2\_m3} & 89.9\% & 1385/1540 & [88.33, 91.34] & 3.41\,s \\
C55 & \texttt{model2vec\_256d} & 256 & \texttt{mxbai\_large} & 89.9\% & 1384/1540 & [88.26, 91.28] & 2.25\,s \\
C56 & \texttt{model2vec\_256d} & 256 & \texttt{no\_reranker} & 89.9\% & 1384/1540 & [88.26, 91.28] & 1.61\,s \\
\bottomrule
\end{tabular}
\end{center}
\end{document}